\documentclass[runningheads]{llncs}

\usepackage{eccv}

\makeatletter
\DeclareRobustCommand\onedot{\futurelet\@let@token\@onedot}
\def\@onedot{\ifx\@let@token.\else.\null\fi\xspace}

\def\eg{\emph{e.g}\onedot} 
\def\ie{\emph{i.e}\onedot}

\makeatother

\def\mG{{\mathcal G}}

\def\mV{{\mathcal V}}

\def\0{{\bf 0}}
\def\1{{\bf 1}}

\def\bF{{\bf F}}
\def\bG{{\bf G}}

\def\bL{{\bf L}}

\def\bP{{\bf P}}

\def\bV{{\bf V}}

\def\bX{{\bf X}}

\def\bZ{{\bf{Z}}}

\def\bp{{\bf p}}

\def\bx{{\bf x}}

\def\bz{{\bf z}}

\def\mmP{{\mathrm P}}

\def\mmR{{\mathbb R}}
\def\mmP{{\mathbb P}}
\def\mmQ{{\mathbb Q}}

\def\citep{\cite}
\def\citet{\cite}

 \makeatletter
\def\@fnsymbol#1{\ensuremath{\ifcase#1\or \dagger\or \ddagger\or
   \mathsection\or \mathparagraph\or \|\or **\or \dagger\dagger
   \or \ddagger\ddagger \else\@ctrerr\fi}}
\makeatother

\usepackage{eccvabbrv}

\usepackage{graphicx}
\usepackage{booktabs}
\usepackage{pifont}
\usepackage{multirow}
\newcommand{\cmark}{\ding{51}}%
\newcommand{\xmark}{\ding{55}}%
\usepackage[accsupp]{axessibility}  %

\usepackage{hyperref}
\definecolor{citecolor}{HTML}{0071bc}
\definecolor{paleplum}{rgb}{0.8, 0.6, 0.8}
\hypersetup{
  colorlinks,
  citecolor=citecolor,
  linkcolor=red
}
\newcommand{\abbrmodel}{LongVLM\xspace}

\usepackage{orcidlink}

\begin{document}

\title{LongVLM: Efficient Long Video Understanding via Large Language Models} 

\titlerunning{\abbrmodel}
\authorrunning{Y. Weng et al.}

\author{Yuetian Weng\inst{1} %
    \and Mingfei Han\inst{3,4} %
    \and Haoyu He\inst{1}  %
    \and \\ Xiaojun Chang\inst{2,3} %
    \and Bohan Zhuang\inst{1}\thanks{Corresponding author.} \\ %
}

\institute{$^1$ZIP Lab, Monash University, Australia
    \quad $^2$School of Information Science and Technology, University of Science and Technology of China \\
    \quad $^3$Department of Computer Vision, MBZUAI 
    \quad $^4$ReLER, AAII, UTS \\
\email{\{yuetian.weng,haoyu.he\}@monash.edu, hmf282@gmail.com, xjchang@ustc.edu.cn, bohan.zhuang@gmail.com}}

\maketitle

\begin{abstract}
Empowered by Large Language Models (LLMs), recent advancements in Video-based LLMs (VideoLLMs) have driven progress in various video understanding tasks. These models encode video representations through pooling or query aggregation over a vast number of visual tokens, making computational and memory costs affordable. Despite successfully providing an overall comprehension of video content, existing VideoLLMs still face challenges in achieving detailed understanding due to overlooking local information in long-term videos. To tackle this challenge, we introduce LongVLM, a simple yet powerful VideoLLM for long video understanding, building upon the observation that long videos often consist of sequential key events, complex actions, and camera movements. Our approach proposes to decompose long videos into multiple short-term segments and encode local features for each segment via a hierarchical token merging module. These features are concatenated in temporal order to maintain the storyline across sequential short-term segments. Additionally, we propose to integrate global semantics into each local feature to enhance context understanding. In this way, we encode video representations that incorporate both local and global information, enabling the LLM to generate comprehensive responses for long-term videos. Experimental results on the VideoChatGPT benchmark and zero-shot video question-answering datasets demonstrate the superior capabilities of our model over the previous state-of-the-art methods. Qualitative examples show that our model produces more precise responses for long video understanding. Code is available at \url{https://github.com/ziplab/LongVLM}. 
\end{abstract}

\section{Introduction}
\label{sec:intro}

\begin{figure}[t]
     \centering
     \begin{subfigure}[b]{0.227\textwidth}
         \centering
         \includegraphics[width=\textwidth]{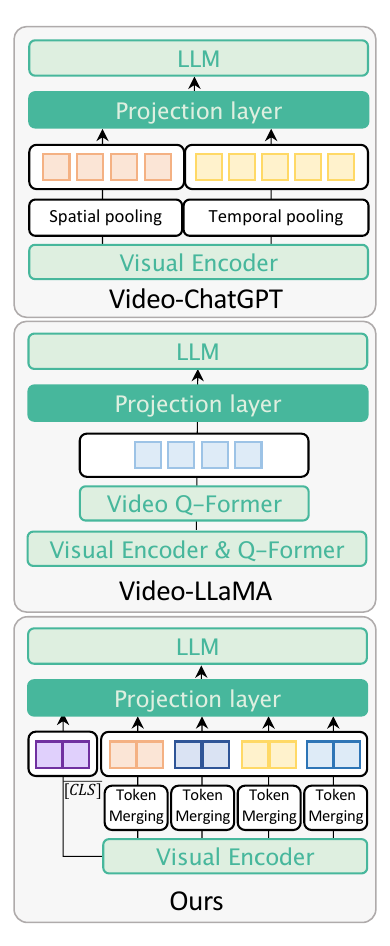}
         \caption{Architectures. 
         }
         \label{fig:arch_comp_l}
     \end{subfigure}
     \hfill
     \begin{subfigure}[b]{0.76\textwidth}
         \centering
         \includegraphics[width=\textwidth]{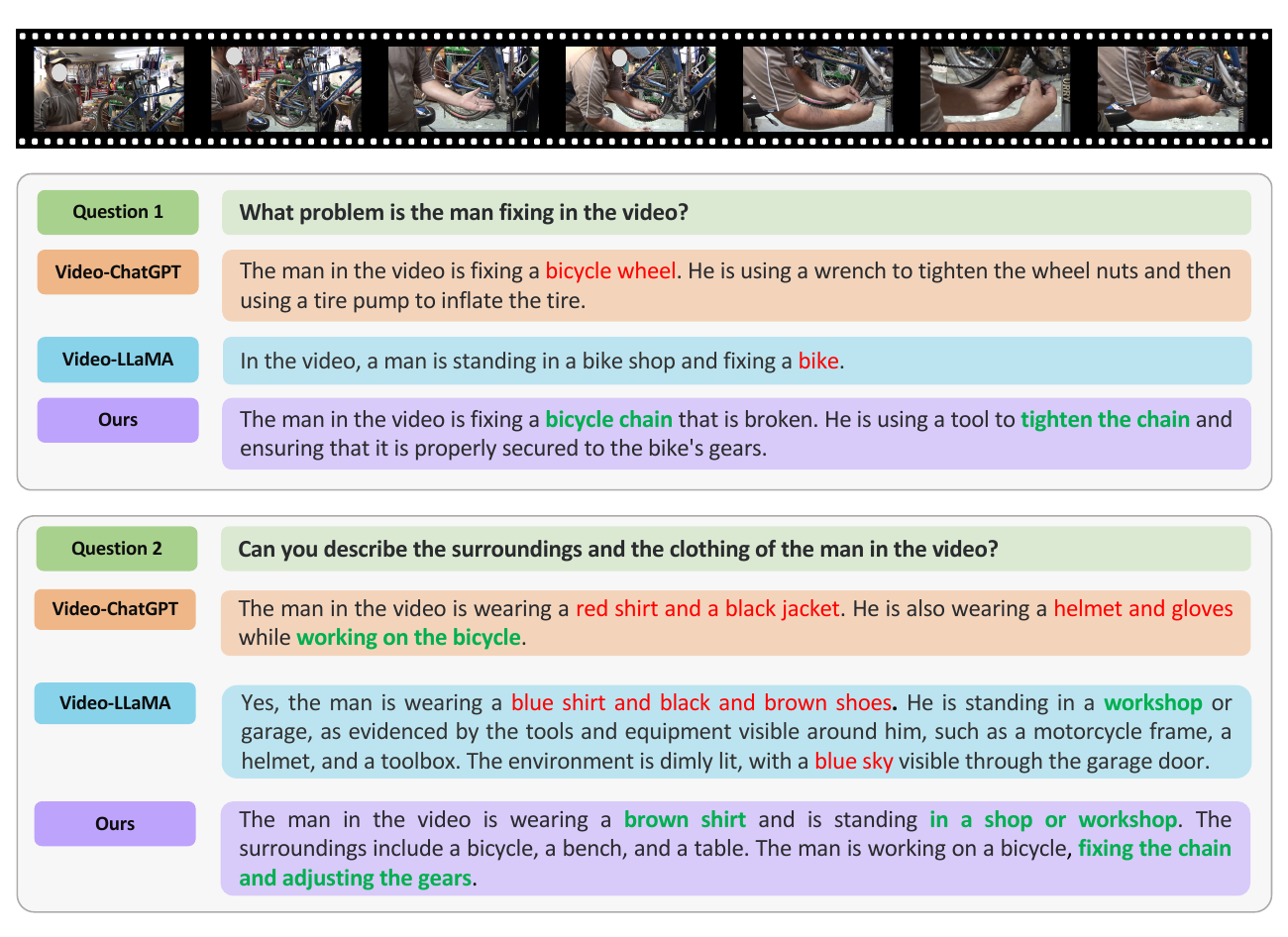}
         \caption{An example from Video-ChatGPT benchmark~\cite{maaz2023video}.}
         \label{fig:motiva_exp}
     \end{subfigure}
     \caption{(a) Comparison of model architectures. (b) Examples generated by different VideoLLMs. Text highlighted in bold green denotes correct content, while text in red indicates errors.}
     \label{fig:arch_compare}
\end{figure}

Large language models (LLMs)~\cite{chatgpt,achiam2023gpt,touvron2023llama,touvron2023llama2,taori2023stanford,chiang2023vicuna} have revolutionized natural language understanding tasks and have demonstrated a remarkable capability to follow human instructions and intentions, emerging as a universal agent for general-purpose assistants. Drawing from the development of LLMs, Multi-modal Large Language Models (MLLMs)~\cite{liu2023visual,zhu2023minigpt,instructblip,gao2023llama} have driven advancements in vision-language learning by integrating visual encoders with LLMs and finetuning on vision-language instruction pairs. However, developing Video-based Large Language Models (VideoLLMs) still poses a significant challenge due to the necessity of processing a large number of tokens for jointly modeling spatial-temporal dependencies across consecutive video frames. For instance, employing OpenAI CLIP-ViT-L/14~\cite{radford2021learning} as a visual encoder for a 100-frame video clip necessitates handling 25.6K visual tokens, leading to impractical computational costs with existing LLMs. 
To address this issue, recent approaches propose to extract video representation via precompression over visual tokens, utilizing pooling operation~\cite{maaz2023video,luo2023valley} or query aggregation~\cite{li2023videochat,zhang2023video,song2024moviechat} over the video token sequence before feeding them into the LLM, as shown in Fig.~\ref{fig:arch_comp_l}. While these models showcase impressive capabilities in providing a meaningful understanding of video content, they still face challenges in achieving significant advantages in fine-grained understanding of long-term videos. For example, as shown in Fig.~\ref{fig:motiva_exp}, while all models recognize the overall environment (\textit{workshop}), the object (\textit{bike}), and the action (\textit{fixing}), previous methods may fail to correctly identify details such as the color of the clothes (\textit{brown}), or the specific component being fixed (\textit{bicycle chain}). 

The main reason is that long-term videos typically involve numerous key actions, complex activities, and camera movements. 
Consequently, a long video can be divided into a sequence of short-term segments. 
For instance, in the example depicted in Fig.~\ref{fig:motiva_exp}, various short-term actions occur, \eg. speaking, displaying spare components, grabbing the bicycle chain, along with the camera moving from the human to the bicycle wheel, and eventually focusing on the broken chain. 
Similarly, prior methods in video recognition task suggest to decompose complex activities into sequences of sub-activities~\cite{wang2013latent,hussein2019timeception,gaidon2013temporal}. These approaches treat the features of each short-range activity as the local information within the videos and emphasize the importance of reasoning over local features to develop a temporal-structural understanding~\cite{wang2013latent,gaidon2013temporal,wang2016temporal,wang2018temporal,hussein2019timeception,zhang2021temporal} within long-term videos for comprehending fine-grained information. 
From this perspective, existing VideoLLMs treat all visual tokens equally and aggregate them into compact representations through pooling operations~\cite{maaz2023video,luo2023valley} and query aggregation~\cite{li2023videochat,zhang2023video}. While they successfully capture the global semantic context spanning the entire long-term videos, they often overlook preserving the local information for the short-term segments and the temporal structure of different short-term components, \eg, the order of events or sub-actions. However, exclusively modeling the temporal structure through the sequence of local features may still lead to inconsistent recognition across different segments and impede the overall understanding of the videos. To comprehend the content in long videos, the human visual system relies on a blend of local and global information~\cite{tian2023view}. 
Building on this insight, earlier approaches in video object detection~\cite{wu2019long, wu2022memvit} suggest integrating global semantics into local localization descriptors, motivating us to include global semantic information into the sequence of local features for enriching the context understanding for each short-term segment. 

In this paper, we present \abbrmodel, a simple yet effective VideoLLM for efficient long video understanding, as illustrated in Fig.~\ref{fig:architecture}. We propose to extract video representations as sequences of short-term local features, and integrate global semantics into each short-term segment feature. 
Specifically, we begin by uniformly sampling a sequence of video frames from long-term videos and utilize a pretrained visual encoder, \eg, CLIP-ViT-L/14~\cite{radford2021learning}, to extract visual features for each individual video frame. 
These frame-level features include the [CLS] tokens from a range of encoder layers and the patch features from the last second layer of the visual encoder. 
Then, we divide the sequence of patch features along the temporal dimension, resulting in multiple short-term segments. Each segment is considered as a local unit in the videos and includes patch features of the video frames within that segment. 
To reduce computational costs and obtain the compact features for each segment, a token merging module is employed to aggregate these patch features for the specific segment into a condensed set of tokens. In this way, we obtain the local features for each segment. We next concatenate these features sequentially to explicitly preserve the temporal order of the short-term segments in long-term videos. Moreover, we average the [CLS] tokens from each video frame along the temporal dimension to represent the global semantic information of the entire video. To integrate the global information, we prepend the averaged [CLS] tokens before the segment-level features, and then feed them into the LLM after passing through a projection layer. Benefiting from the causal attention mechanism in the LLM, we simultaneously achieve temporal structure modeling over the sequence of short-term segments and inject global semantics into the local features. Finally, the LLM generates responses based on the input sequence, which is composed of the obtained video representation and the designed system command with the specific user queries. 

Overall, our main contributions is threefold: 

\begin{itemize}
    \item We propose \abbrmodel, a simple yet effective VideoLLM for efficient long-term video understanding at a fine-grained level while maintaining affordable computational cost. 
    \item We propose decomposing long videos into short segments and extracting local features for each segment to preserve their temporal order. To compactly represent each segment, we propose a hierarchical token merging module to aggregate visual tokens. Additionally, we integrate global semantics into each segment to enhance context understanding. 
    \item Extensive experiments on  VideoChatGPT benchmark~\cite{maaz2023video} and zero-shot video question-answering datasets~\cite{yu2019activitynet,xu2017video} demonstrate that our \abbrmodel surpasses the previous state-of-the-art methods by a significant margin while generating more precise and accurate response at fine-grained level for long-term videos. 
\end{itemize}

\section{Related Work}
\label{sec:related_work}

\subsection{Large Language Models}
Large Language Models (LLMs) have revolutionized natural language processing in recent years. Pretrained on large text corpora, LLMs like GPT~\cite{brown2020language}, OPT~\cite{zhang2022opt}, and LLaMA~\cite{touvron2023llama,touvron2023llama2} utilize auto-regressive Transformer models to predict subsequent tokens, showcasing remarkable adaptability and generalization. Models such as InstructGPT~\cite{ouyang2022training}, ChatGPT~\cite{chatgpt}, and GPT-4~\cite{achiam2023gpt} benefit from instruction-tuning techeque~\cite{wei2021finetuned} on instructional datasets, leveraging the knowledge of pretrained LLMs and demonstrating improvements in diverse conversational interaction capabilities. This strategy is widely adopted in open-source models like Alpaca~\cite{taori2023stanford} and Vicuna~\cite{chiang2023vicuna}, which build upon the advancements made by LLaMA~\cite{touvron2023llama} using specially designed instruction pairs.
Drawing from the advancement of LLMs, recent Multi-modal Large Language Models (MLLMs), \eg, BLIP-2~\cite{li2023blip}, Mini-GPT4~\cite{zhu2023minigpt}, LLaVA~\cite{liu2023visual}, LLama Adapter v2~\cite{gao2023llama}, have demonstrated the feasibility of enabling visual conversation capabilities of LLMs over input images through instruction tuning on image-text instruction datasets. 
Our model aims to utilize existing MLLMs to develop efficient video dialogue model for long-term video understanding. 

\subsection{Video-based Large Language Models}
Traditional video-language models~\cite{luo2020univl,fu2021violet,fang2021clip2video,lei2021less,wang2022internvideo,li2023lavender,cheng2023vindlu,li2023unmasked,han2023html} have advanced by using large-scale video-text pretraining followed by fine-tuning on specific downstream tasks. With the advent of LLMs, Video-based Large Language Models (VideoLLMs) explore various video-language understanding scenarios through human-video dialogue interactions. 
Existing VideoLLMs typically follow a common paradigm, which involves using a pretrained visual encoder to encode visual features, a projection layer to convert visual representations into the text latent space of LLMs, and a pretrained LLM for response generation. VideoChatGPT~\cite{maaz2023video} and Valley~\cite{luo2023valley} rely on pooling over visual tokens to obtain compact visual representations. VideoChat~\cite{li2023videochat} utilizes pretrained video foundation models~\cite{li2023uniformerv2, wang2022internvideo} and Q-Former from BLIP-2~\cite{li2023blip} to aggregate video representations. Video-LLaMA~\cite{zhang2023video} proposes a Video Q-Former and an Audio Q-Former, enabling multiple modalities for video comprehension, while Video-ChatCaptioner~\cite{chen2023video} employs ChatGPT~\cite{chatgpt} to summarize video descriptions in multiple rounds of interactive question-and-answer conversation. Recently, MovieChat~\cite{song2024moviechat} proposes an effective memory management mechanism to enable LLMs to reason over hour-long videos. Multiple video-centric instruction datasets~\cite{luo2023valley,li2023videochat,maaz2023video,han2023shot2story20k} have also been proposed to finetune VideoLLMs for better video understanding capacity. Moreover, BT-Adapter~\cite{liu2023one} proposes a temporal adapter alongside the visual encoder for post-pretraining, while Video-Teller~\cite{liu2023video} highlights the importance of modality alignment in pretraining. 
Overall, these VideoLLMs rely on pooling and query aggregation on the whole long videos to extract visual representation, overlooking local information for fine-grained understanding in long videos. In contrast, we propose a simple yet effective framework that is feasible for aggregating both local and global information in long-term videos and preserves fine-grained content understanding. 

\subsection{Long-term Video Processing}
Long-term video understanding poses several challenges due to the need to exploit complicated spatial-temporal dependencies while removing temporal redundancy over extended time duration. 
Previous studies propose efficient architectures~\cite{donahue2015long,kondratyuk2021movinets}, temporal pooling/aggregation~\cite{fernando2016rank,sener2020temporal,zhou2018temporal,wu2021towards}, dynamic clip selection~\cite{korbar2019scsampler,gowda2021smart,ghodrati2021frameexit} to aggregate video representation while removing redundant information in videos. Other methods in video-language understanding tasks suggest to capture event temporality, causality, and dynamics in long-term videos by designing temporal alignment modules~\cite{han2022temporal,buch2022revisiting}. 
Memory mechanism is also widely adopted in video dense prediction tasks~\cite{lu2017online,wu2019long,yang2018learning,wu2022memvit,zhao2023streaming} to capture historical information and maintain temporal coherence, which results in more accurate and consistent prediction over time in long-term videos. Differently, we propose to aggregate both local segment-level information and global semantic information, empowering MLLMs enhanced fine-grained understanding for long-term videos. 
\newcommand{\videoinput}{$\bV$\xspace}
\newcommand{\localfeature}{\bF_L\xspace}
\newcommand{\globalfeature}{\bF_G\xspace}

\section{Method}
\label{sec:method}

\begin{figure}[t]
  \centering
    \includegraphics[width=1.0\linewidth]{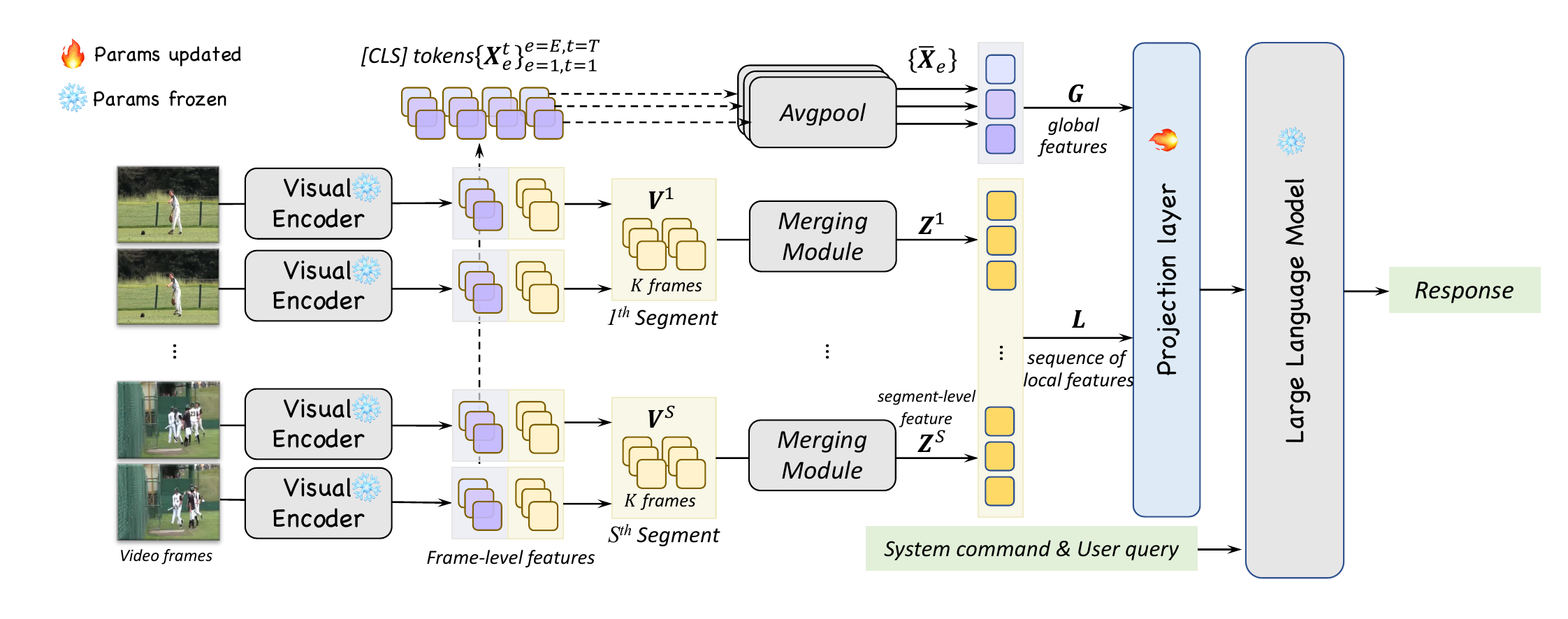}
  \caption{Overall architecture of the proposed \abbrmodel. We start by uniformly sampling $T$ frames from a video and employing a visual encoder to extract frame-level features. We divide the input video into $S$ segments, each with $K$ frames. To obtain compact local features, we apply a hierarchical token merging module within each segment. These segment-level features are concatenated sequentially to explicitly preserve the temporal order of multiple short-term segments in long videos. Additionally, we incorporate [CLS] tokens to aggregate global semantic features. The global features and the sequence of local features are concatenated to form the video representations. Finally, the projected visual features are combined with the tokenized system command and user queries and inputted into the LLM to generate the responses. 
  }
  \label{fig:architecture}
\end{figure}

In Sec.~\ref{sec:overall}, we introduce the overall architecture and generation pipeline of the proposed \abbrmodel. In Sec.~\ref{sec:short_term}, we introduce the process of constructing local representation via short-term feature aggregation. 
In Sec.~\ref{sec:local_global}, we discuss the integration of both local segment-level feature and global semantic feature. 

\subsection{Overall Architecture}
\label{sec:overall}
The overall architecture consists of three components: a visual encoder, a projection layer, and a large language model, as illustrated in Fig.~\ref{fig:architecture}. 

Given an input video $\mV\in\mmR^{T\times H\times W\times3}$, we employ a visual encoder to extract frame-level features $\{\bX^t, \bP^t\}_{t=1}^T$ for each video frame independently. Following previous methods~\cite{liu2023visual,maaz2023video}, we utilize the patch feature $\bP^t\in\mmR^{N\times d}$ from the second-to-last encoder layer, where $N,d$ are the number of patch tokens and the channel dimension of the visual encoder, respectively. Additionally, we gather the [CLS] tokens $\bX^t\in\mmR^{E\times d}$ from $E$ selected encoder layers for each individual video frame. 

To enable fine-grained understanding in long videos, we propose to divide long videos into a sequence of short-term segments, where each segment corresponds to the local features in the long videos. Without loss of generality, the input video $\mV$ is divided into $S$ segments, where each segment includes $K$ frames, \ie, $K=\frac{T}{S}$. 
We collect patch features within the $s^{th}$ segment, \ie, $\bV^s=\{\bP^t\}_{t=(s-1)K}^{t=sK}$, and apply a token merging module $\mG(\cdot)$ to aggregate $\bV^s$ into the compact segment-level feature $\bZ^s=\mG(\bV^s)$. 
These segment-level features are sequentially concatenated as the sequence of local representation $\bL$, explicitly preserving temporal order of short-term segments in long videos. Furthermore, to integrate global semantic information, we propose to collect the [CLS] tokens for each frame from $E$ encoder layers and average them in the time dimension, resulting in our global feature $\bG$. 

We forward the global features and the sequence of local features into a linear layer to obtain the projected visual features. The projected visual features are concatenated with the tokenized system command and user queries, which are inputted into LLM for response generation.

\subsection{Local Feature Aggregation}
\label{sec:short_term}

After obtaining the frame-level patch feature $\{\bP^t\}_{t=1}^{T}$ from the last second layer of visual encoder, previous methods either apply factorized spatio-temporal pooling~\cite{maaz2023video,luo2023valley}, or utilize query aggregation~\cite{li2023videochat,zhang2023video} over all visual tokens, which may miss local information referring to the short-term events or actions. 
Nevertheless, videos have heavy spatio-temporal redundancy, which results in redundant computational costs by directly considering all the patch features as the local representation for each segment. 
Therefore, we propose to aggregate compact visual features within each short-term segment. %
Specifically, we collect the patch feature for the $s^{th}$ segment $\bV^s=\{\bP^t\}_{t=(s-1)K}^{sK}\in\mmR^{KN \times d}$ and apply a hierarchical token merging module to aggregate the local feature while reducing the number of visual tokens. 
Inspired by ToMe~\cite{bolya2022token}, we resort to the bipartite soft matching method and gradually merge the visual tokens for each short-term segment. 
At the $i^{th}$ step, we randomly partition the $R_i$ tokens into two non-overlap token sets $\mmP_i$ with $r_i$ tokens and $\mmQ_i$ with $R_i - r_i$ tokens, where initial $R_0=KN$. 
Then we calculate the similarity scores between the tokens in set $\mmP_i$ and $\mmQ_i$ based on the patch features. To obtain the similarity scores, each visual token is divided into $C$ heads along channel dimension, each with $\frac{d}{C}$ channels. 
The similarity score for each token pair is obtained by averaging the cosine similarity scores over all heads following Eq.~\ref{eq:1}:

\begin{equation} \label{eq:1}
a^{p_{i}q_{i}}=\frac{1}{C}[\sum_{c=1}^{C}{cos(\bp_c^{(p_i)}, \bp_{c}^{(q_i)})}],
\end{equation}
where $p_i\in \{1,..., r_i\}$ and $q_i\in \{1,...,(R_i-r_i)\}$ are the indexes of patch feature $\bp$ in set $\mmP_i$ and set $\mmQ_i$, respectively. 
We select the top-$r_i$ token pairs with the highest similarity scores and merge the paired tokens by average pooling. 
Finally, the remaining tokens in the two sets are concatenated back together, resulting in $R_i-r_i$ tokens after the $i^{th}$ merging step. We iteratively merge the tokens within each short-term segment, until the number of visual tokens reaches $M$, where $M<<K\times N$. The compact local feature for the $s^{th}$ segment is denoted as $\bZ^s=\{\bz_m\}_{m=1}^{M}\in \mmR^{M\times d}$. 
These segment-level features are concatenated sequentially as the sequence of local features $\bL=\{\bZ^s\}_{s=1}^S=[\bz_1^1,...,\bz_M^1,...,\bz_1^S,...,\bz_M^S]\in\mmR^{MS\times d}$. 
Thanks to the positional encoding in the LLM, the sequence of local representation $\bL$ explicitly preserves the order of short-term segments in long-term videos, enabling LLMs to be aware of the temporal structure of multiple event occurrences in long videos. By utilizing the token merging module, we efficiently encode compact local features for each segment while eliminating redundancy in the visual token sequence. %

\subsection{Global Semantics Integration}
\label{sec:local_global}

The local features provide fine-grained information about different events or actions in the generation process of VideoLLMs, enhancing the detailed understanding capabilities of the \abbrmodel. However, the local features for each segment may be insufficient for the model to reason the relationship between different segments and generate reasonable response over the entire videos. 
Therefore, we additionally introduce global semantic features to enrich the local features with contextual information. 
Specifically, we collect the [CLS] tokens  of each video frame from $E$ encoder layers, \ie, $\{\bx^t_e\}_{e=1,t=1}^{e=E,t=T}$, and then average the [CLS] tokens along temporal dimension, resulting in $\bar{\bX}_e=\operatorname{AvgPool}(\{\bx_e^t\}_{t=1}^{T})\in\mmR^{d},e\in[1,...,E]$. 
By default, $E$ can be the number of layers in the visual encoder. However, previous studies demonstrate different properties between intermediate features from the shallow layers and deeper layers in ViT models~\cite{li2021uniformer,pan2022less}, and showcase the deeper layers tend to aggregate global semantics. Thus, we concatenate the $E$-scale features along sequentially, resulting in the global semantic feature for the entire video, \ie, $\bG=\{\bar{\bX}_e\}_{e=1}^{E}\in\mmR^{E\times d}$. 

Following the previous studies, a projection layer converts the visual features into the language space, and then the visual features are concatenated with the instruction as the input of LLM. 
By utilizing the attention mechanism in the LLM, we can easily enable each token in the local feature to attend to the global semantic feature, thereby achieving straightforward injection of global semantics into the local feature.

\noindent{\textbf{Remark.}} 
To address the risk of overlooking detailed understanding in long-term videos, we propose to divide long videos into multiple short-term segments and aggregate local spatial-temporal representation for each segment and preserving the temporal structure over the sequence of local feature vectors. Moreover, we enrich the local features with context information for better response generation by integrating global semantic information into short-term features. 

\section{Experiments}
\label{sec:exps}

\subsection{Experimental Settings}
\noindent{\textbf{Datasets and evaluation metrics.}} We conduct quantitative evaluations of our model using the VideoChatGPT benchmark~\cite{maaz2023video} to assess its performance in generating text from videos. The benchmark comprises 500 videos sampled from ActivityNet-v1.3 dataset~\cite{caba2015activitynet}, with 2000, 2000, 2000, 500, and 1000 questions in terms of five evaluation aspects: Correctness Information(CI), Detail Orientation(DO), Contextual Understanding(CU), Temporal Understanding(TU) and Consistency(C). Additionally, we evaluate the model on the zero-shot question-answering task using the ANET-QA~\cite{yu2019activitynet} dataset, which contains 8000 QA pairs for 800 videos sampled from ActivityNet-v1.3 dataset~\cite{caba2015activitynet}. The videos range from several seconds to minutes long and cover a wide range of daily human activities. We also utilize MSRVTT-QA~\cite{xu2016msr,xu2017video}(72821 QA pairs for 2990 videos) and MSVD-QA~\cite{chen:acl11,xu2017video}(13157 QA pairs for 520 videos) to evaluate the model performance, derived from publicly available video captioning, MSRVTT~\cite{xu2016msr}, and MRVDC~\cite{chen:acl11}, respectively. 
Following the evaluation protocol outlined in Video-ChatGPT~\cite{maaz2023video}, we employ ChatGPT~\cite{chatgpt} for response evaluation and report the generation quality scores on VideoChatGPT benchmark and the answer accuracy and quality scores of models on zero-shot video QA tasks.

\begin{table}[!t]
\begin{center}
\caption{Comparison with state-of-the-art methods on video conversation benchmark~\cite{maaz2023video} in terms of five evaluation aspects and the average scores across all aspects (Mean). We also report the dataset scale used for finetuning the model.}
\label{tab:benchmark}
\begin{tabular}{l|c|cccccc}
\toprule
\textbf{Method} & \textbf{Data Source} & \textbf{CI}   & \textbf{DO}   & \textbf{CU}   & \textbf{TU}   & \textbf{C}    & \textbf{Mean} \\ \midrule
VideoChat~\cite{li2023videochat}  & 10M & 2.25 & 2.50 & 2.54 & 1.98 & 1.84 & 2.22 \\
LLaMA Adapter v2~\cite{gao2023llama}    & 700K                     & 2.03          & 2.32          & 2.30          & 1.98          & 2.15          & 2.16          \\
Video LLaMA~\cite{zhang2023video}      & 10M                  & 1.96          & 2.18          & 2.16          & 1.82          & 1.79          & 1.98          \\
Video-ChatGPT~\cite{maaz2023video}    & 100K                 & 2.50          & 2.57          & 2.69          & 2.16          & 2.20          & 2.42          \\
Valley~\cite{luo2023valley}  & 234k & 2.43 & 2.13 & 2.86 & 2.04 & 2.45 & 2.38 \\
BT-Adapter~\cite{liu2023one}       & 10M                  & 2.16          & 2.46          & 2.89          & 2.13          & 2.20          & 2.37          \\
BT-Adapter~\cite{liu2023one}       & 10M+100K             & 2.68 & 2.69   & 3.27 & 2.34    & 2.46    & 2.69    \\
\textbf{Ours}   & 100K                 & \textbf{2.76}    & \textbf{2.86} & \textbf{3.34} & \textbf{2.39} & \textbf{3.11} & \textbf{2.89} \\ \bottomrule
\end{tabular}
\end{center}
\end{table}

\begin{table}[!t]
\begin{center}
\caption{Comparison with state-of-the-art methods on three zero-shot question answering datasets. We report the Accuracy (Acc.) and Score for the generated answer for each question, and the dataset scale used for finetuning the model.}
\label{tab:zeroshotqa}

\begin{tabular}{l|c|cc|cc|cc}
\toprule
\multirow{2}{*}{\textbf{Method}} & \multirow{2}{*}{\textbf{Data source}} & \multicolumn{2}{c|}{\textbf{ANET-QA}} & \multicolumn{2}{c|}{\textbf{MSRVTT-QA}} & \multicolumn{2}{c}{\textbf{MSVD-QA}} \\
 &  &  Acc.  & Score & Acc.  & Score & Acc.  & Score \\
\midrule
FrozenBiLM~\cite{yang2022zero}      & 10M                   & 24.7                   & -                  & 16.8                & -                & 32.2               & -               \\
VideoChat~\cite{li2023videochat}       & 10M         & 26.5                   & 2.2                & 45.0                & 2.5              & 56.3               & 2.8             \\
LLaMA Adapter v2~\cite{gao2023llama}   &  700K & 34.2                   & 2.7                & 43.8                & 2.7              & 54.9               & 3.1             \\
Video LLaMA~\cite{zhang2023video}     & 10M         & 12.4                   & 1.1                & 29.6                & 1.8              & 51.6               & 2.5             \\
Video-ChatGPT~\cite{maaz2023video}    & 100K           & 35.2                   & 2.7                & 49.3                & 2.8              & 64.9               & 3.3             \\
Valley~\cite{luo2023valley}  & 234K & 45.1 & 3.2 & 51.1 & 2.9 & 60.5 & 3.3 \\
BT-Adapter~\cite{liu2023one}       & 10M+100K & 45.7                   & 3.2                & 57.0                & 3.2              & 67.5               & 3.7             \\
Ours            & 100K           & \textbf{47.6}          & \textbf{3.3}       & \textbf{59.8}       & \textbf{3.3}     & \textbf{70.0}      & \textbf{3.8}    \\ \bottomrule
\end{tabular}
\end{center}
\end{table}

\noindent\textbf{Implementation details.} 
We employ CLIP-ViT-L/14~\cite{radford2021learning}
as the visual encoder and Vicuna-7B-v1.1~\cite{chiang2023vicuna} as the LLM. We initialize them with the pretrained weights in LLaVA-7B-v1.1~\cite{liu2023visual}. 
We finetune the model on the Video-ChatGPT-100K instruction dataset~\cite{maaz2023video} for 3 epochs, with a learning rate of $2e^{-5}$ and a batch size of 32. We only finetune the linear projection layer to align the visual features into the input space of the LLM, keeping both the visual encoder and LLM frozen. It takes three hours to train three epochs on 4 A100 80GB GPUs. 
During training and inference, we sample $T=100$ video frames for each video, and resize the frames to $224\times 224$ resolutions. We set $C=16$, the same to the number of heads of CLIP-ViT/L-14. 
We set $S=10$ for each video, and the number of tokens in each segment-level feature is $M=30$. We collect the [CLS] tokens from the last five encoder layers and average them along the temporal axis, resulting in $E=5$ tokens as the global semantic features. Therefore, the length of visual tokens for a video sequence is $M\times S + E = 305$.

\subsection{Main Results}
\label{sec:main_results}
\noindent\textbf{Results on the video-based generation benchmark.} In Tab.~\ref{tab:benchmark}, we present a comprehensive evaluation of our \abbrmodel against state-of-the-art models on the video-based generation benchmark~\cite{maaz2023video}. Our \abbrmodel outperforms all other models across all the evaluation aspects. Particularly noteworthy is its significant advantage in Detail Orientation (DO) and Consistency (C),
showing improvements of +0.17 and +0.65, respectively, over BT-Adapter~\cite{liu2023one}. These results underscore superior capability of LongVLM in fine-grained video understanding and robust generation performance. 

\noindent\textbf{Results on zero-shot video question-answering.} In Tab.~\ref{tab:zeroshotqa}, we compare the performance of \abbrmodel against various existing methods on three zero-shot video QA datasets: ANET-QA~\cite{yu2019activitynet,caba2015activitynet}, MSRVTT-QA~\cite{xu2016msr,xu2017video} and MSVD-QA~\cite{chen:acl11,xu2017video}. Our model achieves the highest accuracy of 47.6\%, 59.8\%, and 70.0\% on the three QA datasets, surpassing the previous SOTA approach BT-Adapter~\cite{liu2023one} by 1.9\%, 2.8\% and 2.5\%, respectively. Furthermore, we achieve the highest score in terms of generation quality over the three datasets.

\subsection{Ablation Study}
\noindent\textbf{Effects of local feature aggregation}.
As discussed in Sec.~\ref{sec:intro}, pooling operations or query aggregation might overlook local information in achieving fine-grained understanding in long-term videos. To this end, we introduce short-term segment-level features to retain local information and temporal structure within long-term videos. The first two rows in Tab.~\ref{tab:localglobal} present the effects for the design of using local features as the visual representations for videos. We compare the token merging module with a local pooling operation. Specifically, we apply a 3D average pooling operation within each short-term segment, using a kernel size and stride of (5, 4, 4) to the temporal, height, and width dimensions, respectively. 
The proposed hierarchical merging module achieves higher scores over spatial-temporal pooling operation. This could be attributed to the dynamic aggregation mechanism via the token similarity in the merging module, while averaging pooling statically aggregates visual tokens within each small 3D window. 
Additionally, we observe that aggregating local features for short-term segments either improves or maintains comparable performance across all evaluation metrics compared to the SOTA models which extract global semantics only, highlighting the significance of preserving local features for short-term segments in long-term video understanding.

\begin{table}[!t]
\begin{center}
\caption{Ablation of the local and global aggregation design. Pooling: Using 3D average pooling operation to obtain local features; Merging: Using the proposed hierarchical token merging to obtain local features; $[\bL, \bG]$: Concatenating local feature then global feature; $[\bG, \bL]$: Concatenating global feature then local feature.} 
\label{tab:localglobal}

\begin{tabular}{c|cc|cccccc}
\toprule
\textbf{Variants} & \textbf{Local} & \textbf{Global} & \textbf{CI}   & \textbf{DO}   & \textbf{CU}   & \textbf{TU}   & \textbf{C}    & \textbf{Mean} \\ \midrule
Pooling & \cmark & \xmark & 2.53  & 2.64  & 3.13  & 2.29  & 2.61  & 2.64  \\
Merging & \cmark & \xmark & 2.62  & 2.74  & 3.15  & 2.23  & 2.86  & 2.72  \\
\midrule
$[\bL, \bG]$ & \cmark & \cmark & 2.69 & 2.81 & 3.31 & 2.31 & 2.99 & 2.82 \\
$[\bG, \bL]$ & \cmark & \cmark & \textbf{2.76} & \textbf{2.86} & \textbf{3.34} & \textbf{2.39} & \textbf{3.11} & \textbf{2.89} \\ \bottomrule
\end{tabular}
\end{center}
\end{table}

\begin{table}[!t]
    \caption{Effect of $M$. We evaluate the model performance by varying $M$ from $\{10, 20, 30, 40\}$, while keeping the token length of global semantic features at $E=5$. %
    }
    \label{tab:ablation_toklength}
    \begin{subtable}[h]{0.485\textwidth}
    \centering 
        \begin{tabular}{c|cccccc}
\toprule
\textbf{M} & \textbf{CI} & \textbf{DO} & \textbf{CU} & \textbf{TU} & \textbf{C} & \textbf{Mean} \\ \midrule
\textbf{10}     & 2.61 & 2.72 & 3.22 & 2.26 & 2.78 & 2.72 \\
\textbf{20}      & 2.72 & \textbf{2.86} & 3.34 & 2.34 & 2.96 & 2.84 \\
\textbf{30}      & \textbf{2.76} & \textbf{2.86} & 3.34 & \textbf{2.39} & \textbf{3.11} & \textbf{2.89} \\
\textbf{40}      & 2.74 & 2.84 & \textbf{3.39} & 2.34 & 3.06 & 2.87 \\ \bottomrule
\end{tabular}
        \caption{Video-ChatGPT Benchmark.}
        \label{tab:m_ben}
    \end{subtable}
    \hfill
    \begin{subtable}[h]{0.485\textwidth}
        \begin{tabular}{c|ccc}
\toprule
\textbf{M}  & \textbf{Accuracy} & \textbf{Score} & \textbf{Memory(G)} \\
\midrule
\textbf{10} & 44.6         & 3.2            & \textbf{14.65}                 \\
\textbf{20} & 45.7         & \textbf{3.4}            & 14.74                 \\
\textbf{30} & \textbf{47.6}         & 3.3            & 14.86                 \\
\textbf{40} & 46.0         & 3.3           & 14.96       \\       \bottomrule
\end{tabular}
        \caption{ANET-QA.}
        \label{tab:m_anetqa}
    \end{subtable}
\end{table}

\begin{table}[!t]
\begin{center}
\caption{Effect of the number of selected encoder layers. We evaluate the performance of model varying $E$ in \{1, 5, 10, 15, 20, 24\}, keeping $M=30, K=10$.} 
\label{tab:ablation_global_length}

\begin{tabular}{c|cccccc}
\toprule
\textbf{E}  & \textbf{CI} & \textbf{DO} & \textbf{CU} & \textbf{TU} & \textbf{C} & \textbf{Mean} \\ \midrule
\textbf{1} & 2.74 & 2.85 & 3.23 & 2.32 & 3.04 & 2.83 \\
\textbf{5} & 2.76 & \textbf{2.86} & \textbf{3.34} & \textbf{2.39} & \textbf{3.11} & \textbf{2.89} \\
\textbf{10} & \textbf{2.78} & \textbf{2.86} & 3.24 & 2.30 & 3.04 & 2.84 \\
\textbf{15} & 2.72 & 2.82 & 3.16 & 2.16 & 2.97  & 2.77 \\
\textbf{20} & 2.63 & 2.77 & 3.11 & 2.28 & 2.93 & 2.74 \\ 
\textbf{24} & 2.65 & 2.75 & 3.08 & 2.22 & 2.81 & 2.70 \\ 
\bottomrule
\end{tabular}
\end{center}
\end{table}

\begin{figure}[t]
     \centering
     \begin{subfigure}[b]{0.48\textwidth}
         \centering
         \includegraphics[width=\textwidth]{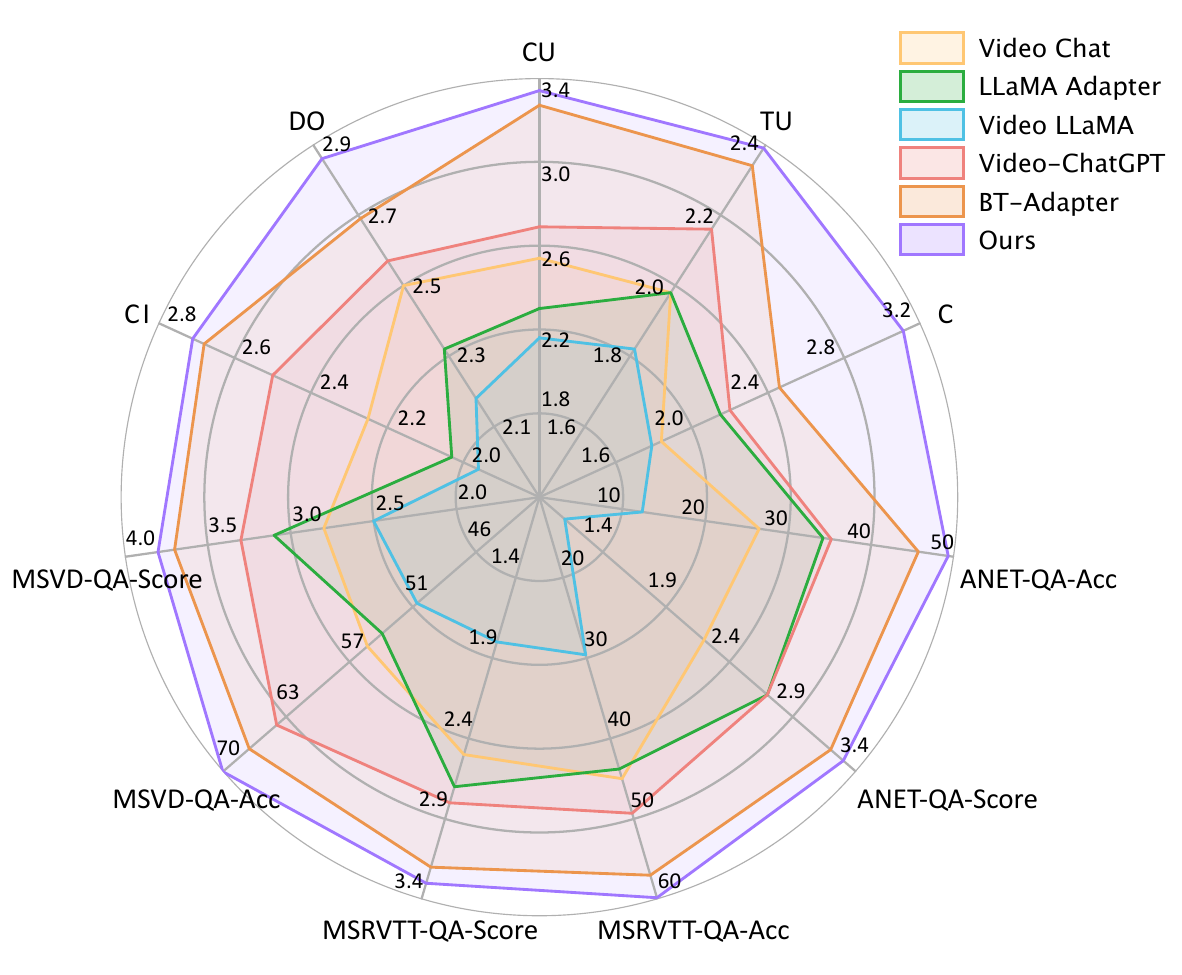}
         \caption{Quantitative results on Video-ChatGPT-100K~\cite{maaz2023video} benchmark and the task of zero-shot question answering on ANet-QA~\cite{yu2019activitynet}, MSRVTT-QA~\cite{xu2016msr,xu2017video} and MSVD-QA~\cite{chen:acl11,xu2017video}. Our model delivers the best performance on multiple evaluation aspects, compared with the state-of-the-art video dialogue models: Video Chat~\cite{li2023videochat}, LLaMA Adapter~\cite{gao2023llama}, Video LLaMA~\cite{zhang2023video}, Video-ChatGPT~\cite{maaz2023video}, and BT-Adapter~\cite{liu2023one}. Evaluation metrics and comparison details are given in Section~\ref{sec:main_results}.
         }
         \label{fig:performance}
     \end{subfigure}
     \hfill
     \begin{subfigure}[b]{0.48\textwidth}
         \centering
         \includegraphics[width=\textwidth]{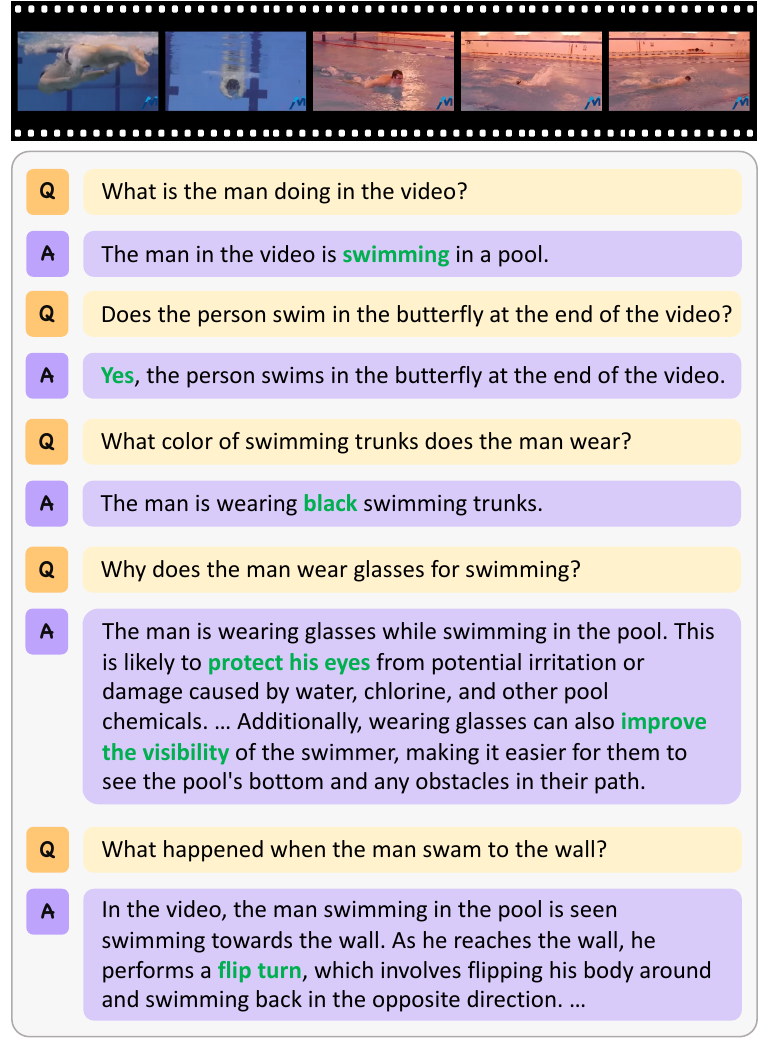}
         \caption{An example of zero-shot question answering. Video duration is 3 minutes and 46 seconds.}
         \label{fig:example_anet}
     \end{subfigure}
     \caption{Quantitative results and qualitative examples of our LongVLM.}
     \label{fig:comparison}
\end{figure}

\begin{figure}[!t]
  \centering
  \includegraphics[width=0.98\linewidth]{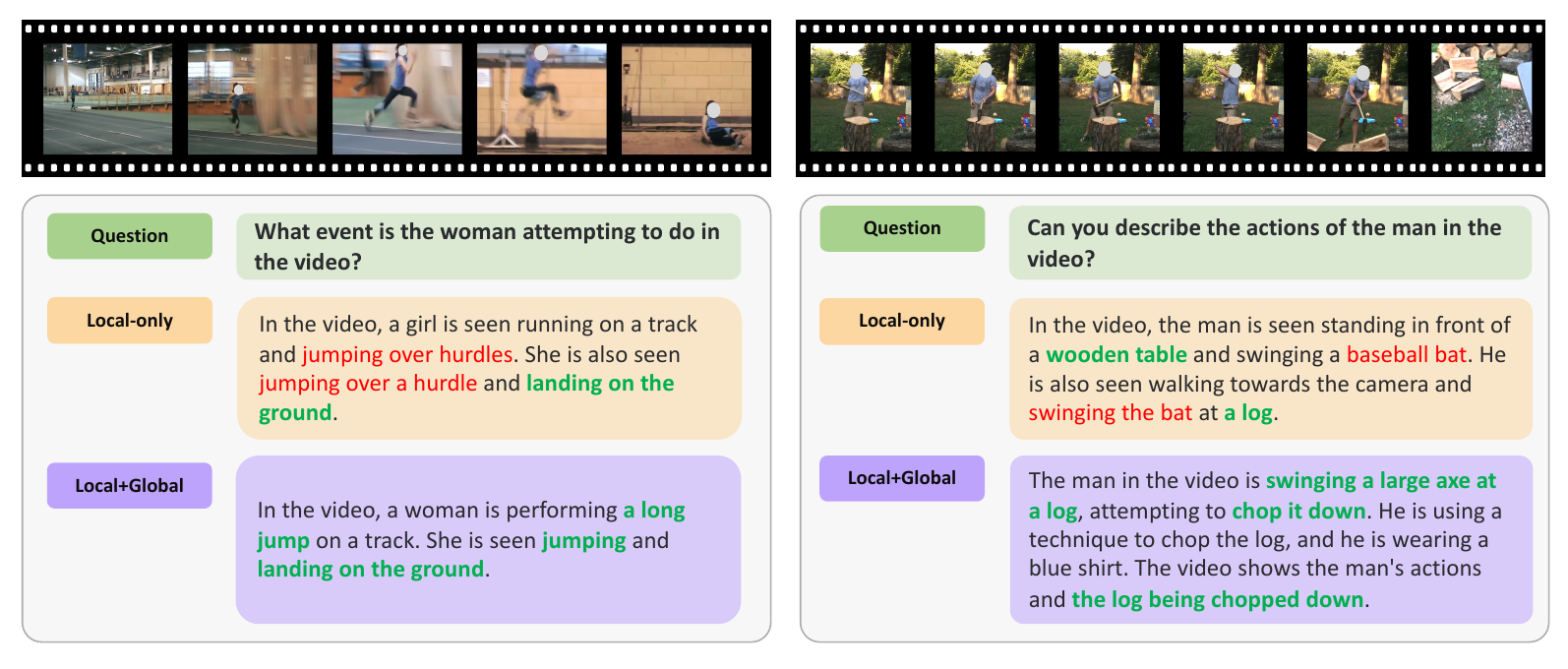}
  \caption{Two examples from Video-ChatGPT benchmark~\cite{maaz2023video}.
  Text highlighted in bold green denotes correct content, while text in red indicates errors. 
  }
  \label{fig:localglobal_example}
\end{figure}

\noindent\textbf{Effects of global semantics integration.} Inspired by the human visual system that using a combination of local and global information for recognizing video content~\cite{tian2023view}, we propose to enhance visual representation by injecting global semantic features into local features. 
The last two rows in Tab.~\ref{tab:localglobal} demonstrate the effects of integrating global semantics. Compared to the first two rows, introducing global semantic features significantly enhances performance compared to models using local features only across all evaluation aspects. The notable improvements in Contextual Understanding (CU) and Consistency (C) underscore the significance of integrating global semantic information with local short-term features. Moreover, concatenating global features before local features yields better results than the opposite concatenation order. This arrangement allows each the local feature to access the global semantic information across the entire video by leveraging the causal attention mechanism in the LLM. Consequently, this design enriches the contextual information of the local features and enhances the response consistency of the model.

\begin{figure}[!t]
  \centering
  \includegraphics[width=0.92\linewidth]{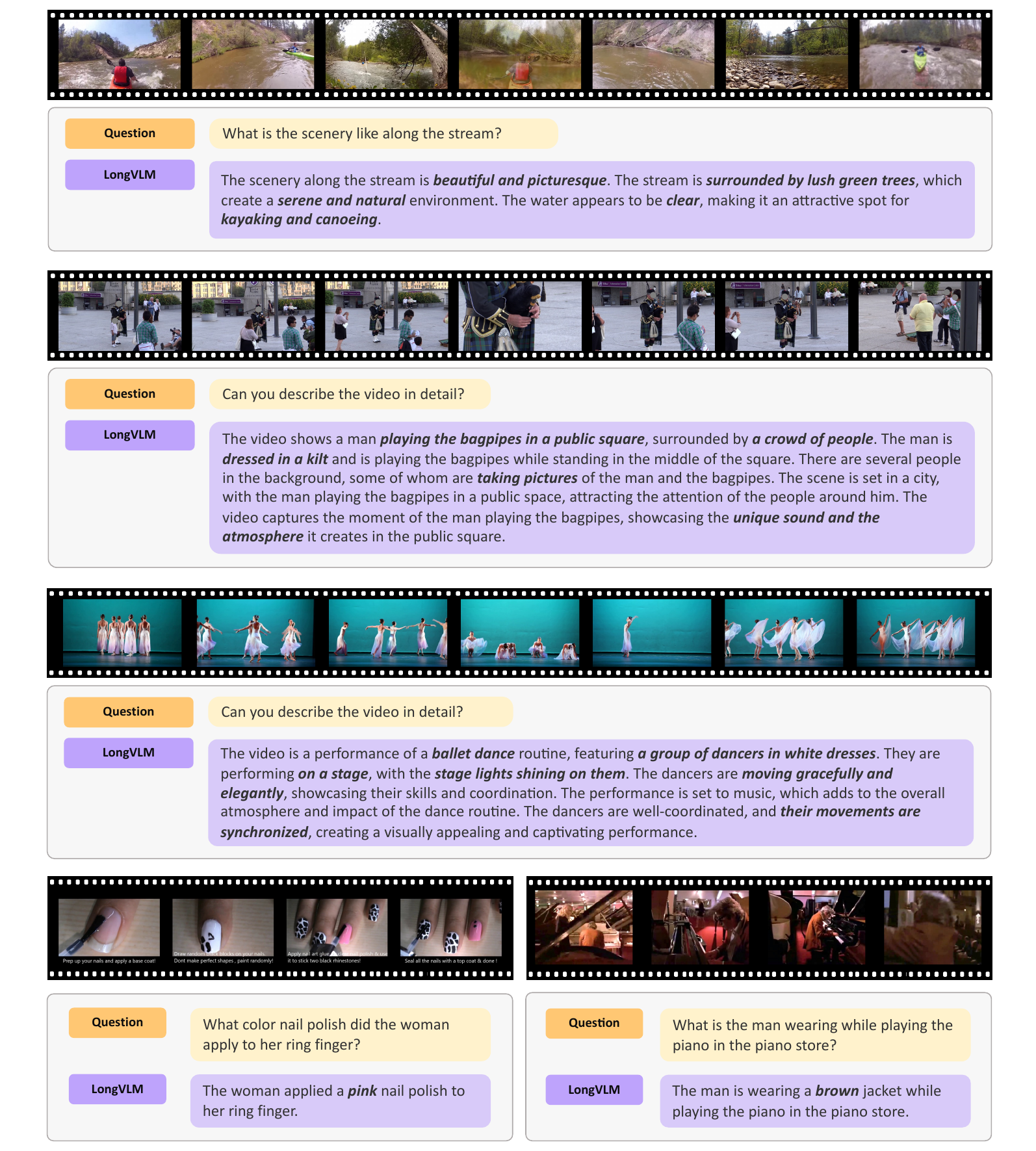}
  \caption{More generative examples from the Video-ChatGPT benchmark~\cite{maaz2023video} of the proposed \abbrmodel. Text in bold denotes the correct content. The \abbrmodel is able to capture the detailed information videos. 
  }
  \label{fig:more_gen}
\end{figure}

\noindent\textbf{Effects of $M$.} We report the model performance on VideoChatGPT benchmark and ANET-QA task on the selection of $M$, \ie, $M=\{10, 20, 30, 40\}$, keeping the same number of global semantic tokens in 
Tab.~\ref{tab:ablation_toklength}. For ANET-QA, we also report the averaging GPU memory usage for generating each answer. In general, the token length involves a trade-off between memory costs and performance. A shorter token sequence reduces computational costs for generating a single new token using LLM, thereby lowering memory costs for generating responses to individual user queries. However, it may also lead to insufficient visual information for generating accurate responses. 
The performance of our model is beneficial from the suitable length of visual tokens. Increasing $M$ from 10 to 40 results in a significant improvement in terms of most evaluation aspects, while the setting of $M=40$ leads to neglecting improvement but requires more memory cost compared to $M=30$. Therefore, we choose $M=30$ for our model.

\noindent\textbf{Effects of $E$.} We evaluate the model performance on the VideoChatGPT benchmark with varying $E$ by selecting the [CLS] tokens from the last $1, 5, 10, 15, 20, 24$ visual encoder layers, while maintaining the same $M$ for local features. As depicted in Tab.~\ref{tab:ablation_global_length}, increasing the number of global semantic tokens from 1 to 5 improves the generation quality scores in terms of all evaluation aspects. However, increasing $E$ from 5 to 24 leads to degraded performance, possibly because the [CLS] tokens from earlier layers carry less semantic information for the model. Therefore, we choose $E=5$ in our model.

\subsection{Qualitative Results}
As illustrated in Sec.~\ref{sec:intro}, Fig.~\ref{fig:motiva_exp} demonstrates the advancement of our model in terms of fine-grained understanding in long-term videos. Despite taking the same number of video frames as input, our model excels in capturing detailed information within the videos, discerning nuances like \textit{fixing chain} rather than \textit{fixing wheel}. In comparison, Video-ChatGPT~\cite{maaz2023video} can describe the overall video content but may inaccurately recognize detailed information. For instance, it might identify objects such as \textit{helmets} and \textit{gloves} in the scene but erroneously recognize the location for these objects. This emphasize the importance of decomposing long videos into multiple short-term segments and aggregate local features to achieve fine-grained understanding in videos. 
The examples depicted in Fig.~\ref{fig:localglobal_example} ablate the effectiveness of integrating global semantic information into local short-term features. 
With global semantics integration, the model is able to recognize the actions (\textit{long jump}) and objects (\textit{axe}) compared to the variant that using local features only. 
We provide more generated examples in Fig.~\ref{fig:example_anet} and Fig.~\ref{fig:more_gen} from ANET-QA and Video-ChatGPT benchmark, respectively, which showcase the precise description of the video content generated by our \abbrmodel. 

\section{Conclusion}
In this work, we have introduced \abbrmodel, an effective and efficient VideoLLM designed for long-term video understanding. 
By extracting local features for short-term segments, we efficiently model local dependencies while preserving the temporal structure of sequential events in long-term video sequences. Through the integration of local and global information, \abbrmodel captures detailed information and provides consistent and accurate responses for long-term videos. 

\noindent\textbf{Limitations and Further work.} 
While we introduce a novel video conversation model for fine-grained long-term video comprehension, our framework is specifically designed for video-to-text generation scenarios. Future work may include extending our framework into video-centric multimodal generation tasks and training the model on large-scale, extended-duration videos for long-context understanding.

\bibliographystyle{splncs04}
\bibliography{reference}

\clearpage

\begin{center}
    \Large{\textbf{Appendix}}
\end{center}

We organize our supplementary material as follows. 
\begin{itemize}
    \item In Sec.~\ref{sec:add_res}, we provide additional ablation results.
    \item In Sec.~\ref{sec:vis_examples}, we provide more generation results from the proposed \abbrmodel. 
\end{itemize}

\section{Additional Results}
\label{sec:add_res}
\begin{figure}[ht]
     \centering
     \begin{subfigure}[b]{0.32\textwidth}
         \centering
         \includegraphics[width=\textwidth]{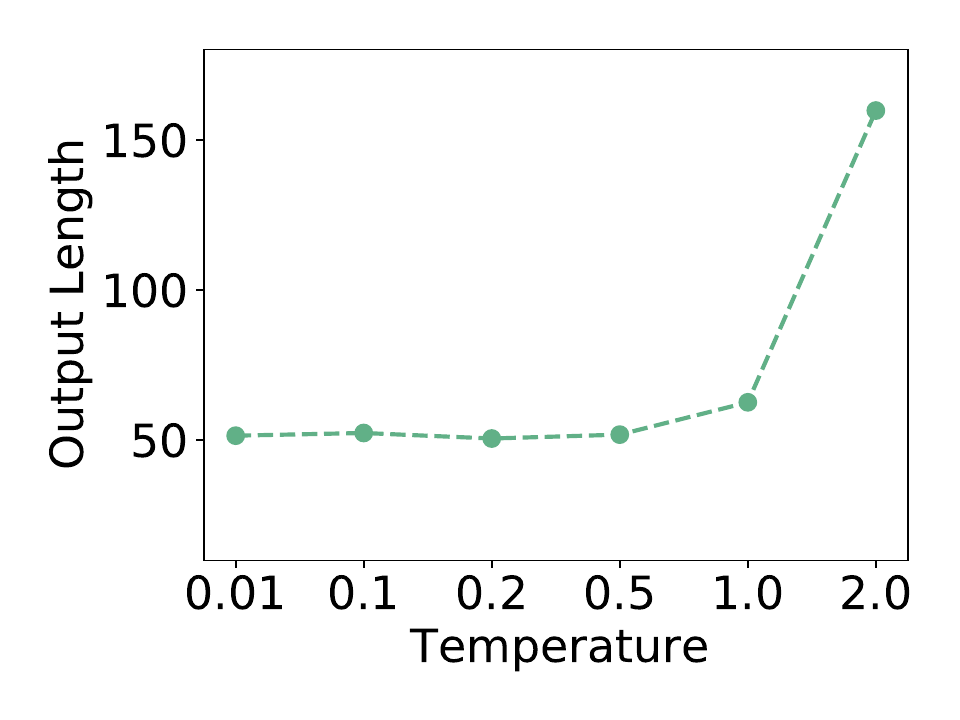}
         \caption{\textbf{CI} 
         }
         \label{fig:ci_length}
     \end{subfigure}
     \hfill
     \begin{subfigure}[b]{0.32\textwidth}
         \centering
         \includegraphics[width=\textwidth]{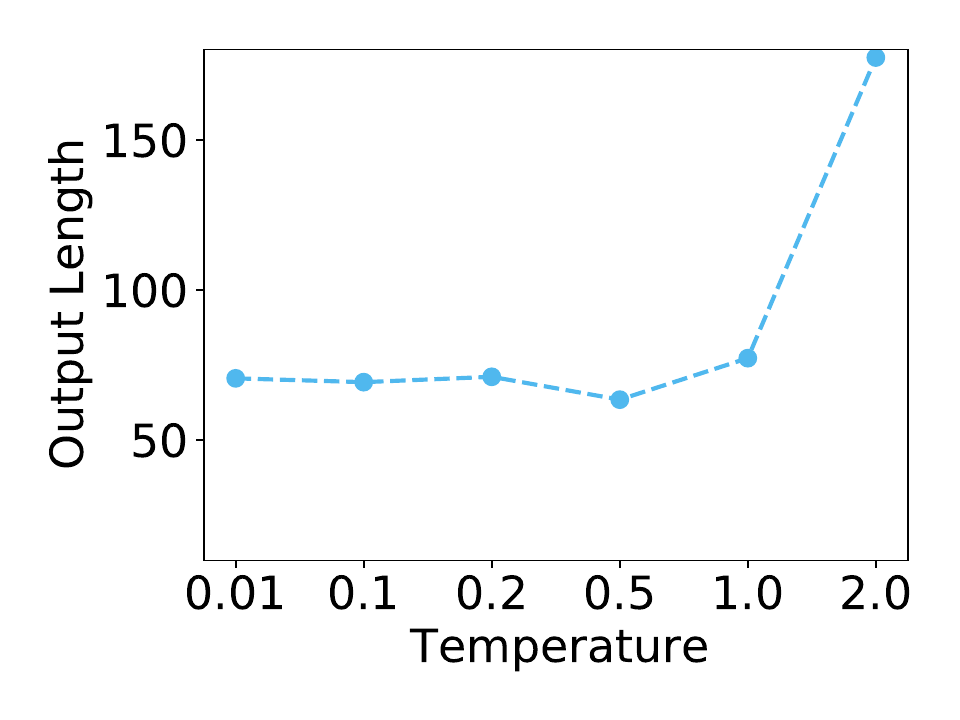}
         \caption{\textbf{TU}}
         \label{fig:tu_length}
     \end{subfigure}
     \hfill
     \begin{subfigure}[b]{0.32\textwidth}
         \centering
         \includegraphics[width=\textwidth]{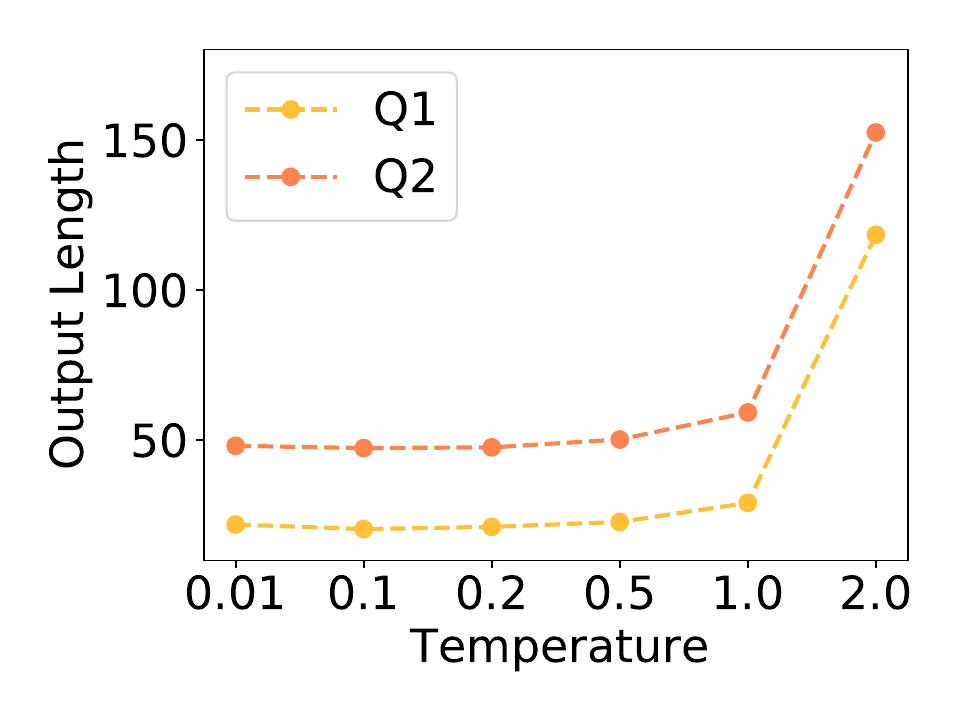}
         \caption{\textbf{C}}
         \label{fig:c_length}
     \end{subfigure}
     \vspace{-0.5em}
     \caption{Average word counts in responses for various temperature settings. "Q1" and "Q2" denote two questions addressing similar perspectives in the Consistency (C) evaluation metrics.}
     \label{fig:output_length}
     \vspace{-0.5em}
\end{figure}

\noindent{\textbf{Effects of temperature.}} We show the effects of temperature settings in the LLM on both performance scores and averaged word counts in responses on Video-ChatGPT benchmark~\cite{maaz2023video}. As demonstrated in Tab.~\ref{tab:temp} and Fig.~\ref{fig:output_length}, lower temperatures lead to shorter but more accurate responses, with tokens that are more deterministic and closely aligned with the most relevant predictions. However, increasing the temperature from 0.2 to 2.0 lowers performance and raises word counts, leading to longer, more creative responses that may be irrelevant and inaccurate. Therefore, we set the temperature at 0.2 in our model. 

\begin{table}[ht]
\begin{center}
\vspace{-1.5em}
\caption{Effects of temperature. We report the Correctness of Information (\textbf{CI}, Temporal Understanding (\textbf{TU}), and Consistency (\textbf{C}) by varying temperature in \{0.01, 0.1, 0.2, 0.5, 1.0, 2.0\}.}
\label{tab:temp}
\begin{tabular}{c|cccccc}
\toprule
\textbf{Temperature} & \textbf{0.01} & \textbf{0.1} & \textbf{0.2} & \textbf{0.5} & \textbf{1.0} & \textbf{2.0} \\ \midrule
\textbf{CI}  & 2.74 & \textbf{2.76} & \textbf{2.76} & 2.69 & 2.35 & 1.21 \\
\textbf{TU}  & 2.22 & 2.16 & \textbf{2.39} & 2.20 & 1.99 & 1.25 \\
\textbf{C}  & 2.85 & 2.89 & \textbf{3.11} & 2.62 & 2.21 & 1.01 \\ \bottomrule
\end{tabular}
\vspace{-2.5em}
\end{center}
\end{table}

\noindent{\textbf{More results on zero-shot QA.}} We evaluate our model in a zero-shot manner on subsets of two datasets: (1) Egoschema~\cite{mangalam2024egoschema}, a long-form video QA dataset derived from Ego4D~\cite{grauman2022ego4d}; and (2) MAD~\cite{soldan2022mad}, an audio description (AD) generation benchmark that requires understanding long contexts in hour-long movies. 
We sample videos at 1 fps for Egoschema and 5 fps for MAD. We report the accuracy (\%) for Egoschema and the response score (ranging from 0 to 5) for MAD in Tab.~\ref{tab:longvideo}, demonstrating the effectiveness of the proposed LongVLM on longer video datasets. 

\begin{table}[!t]
\centering
\caption{Model performance on Egoschema and MAD.}
\label{tab:longvideo}
\begin{tabular}{c|c|c}
\toprule
\textbf{Method}  & \textbf{Egoschema} & \textbf{MAD} \\ \midrule
VideoChat     &  46.6         &  1.90   \\
Video LLaMA   &  38.8         &  1.86   \\
Video-ChatGPT &  49.6         &  1.93   \\
BT-Adapter    &  54.6         &  2.14   \\ 
LongVLM       &  \textbf{57.6}  &  \textbf{2.21} \\ \bottomrule
\end{tabular}
\end{table}

\noindent\textbf{Additional ablation results on zero-shot QA.} We verify the local feature aggregation and global semantics integration strategies on ANET-QA~\cite{yu2019activitynet} and Egoschema~\cite{mangalam2024egoschema}. We report the accuracy (\%) in Tab.~\ref{tab:add_abla}. The results demonstrate the effectiveness of token merging module compared to the pooling operation on longer videos. Additionally, concatenating global features before local features yields better results than the opposite concatenation order. 

\begin{table}[!t]
\centering
\caption{Ablation of the local and global aggregation design.}
\label{tab:add_abla}
\begin{tabular}{c|cc}
\toprule
\textbf{Variants}   & \textbf{ANET-QA} & \textbf{Egoschema} \\ \midrule
Pooling    &  45.4       &   48.0        \\
Merging    &  46.5       &   52.4        \\
\midrule
{[}L, G{]} &  47.0       &   56.0        \\
{[}G, L{]} &  \textbf{47.6}  &   \textbf{57.6}  \\ 
\bottomrule
\end{tabular}
\vspace{-1em}
\end{table}

\section{Additional Qualitative Examples}
\label{sec:vis_examples}
We provide more generated examples from VideoChatGPT benchmark, ANET-QA and MSRVTT-QA. The proposed LongVLM can generate precise and accurate responses that correspond to video content and correctly answer questions about detailed information in videos.

\begin{figure}[!t]
  \centering
  \includegraphics[width=1.0\linewidth]{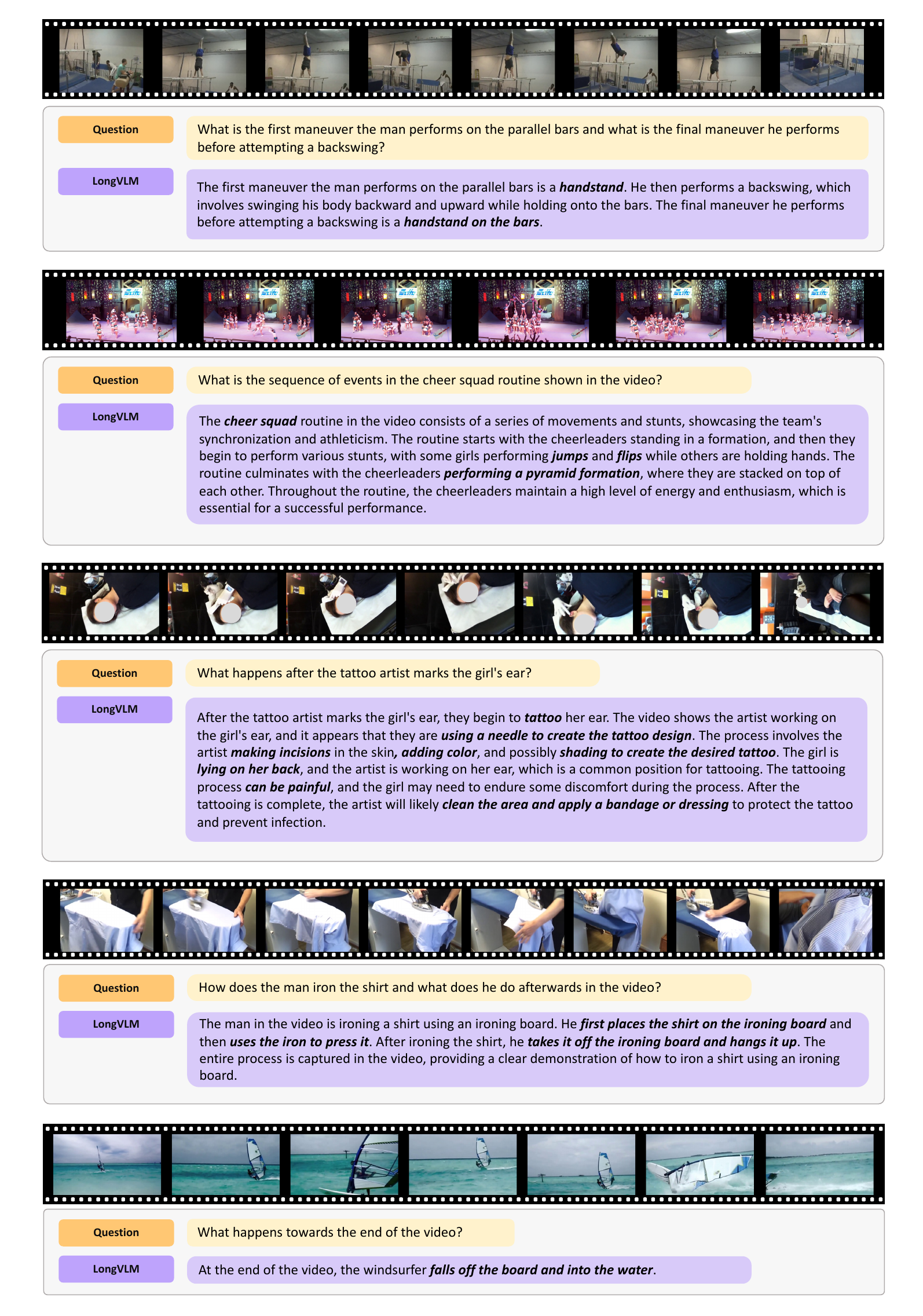}
  \caption{Additional response examples from the Video-ChatGPT benchmark~\cite{maaz2023video} generated by our \abbrmodel, focusing on temporal understanding (\textbf{TU}) evaluation perspective. Text in bold indicates the correct content.}
  \label{fig:more_tem}
\end{figure}

\begin{figure}[!t]
  \centering
  \includegraphics[width=1.0\linewidth]{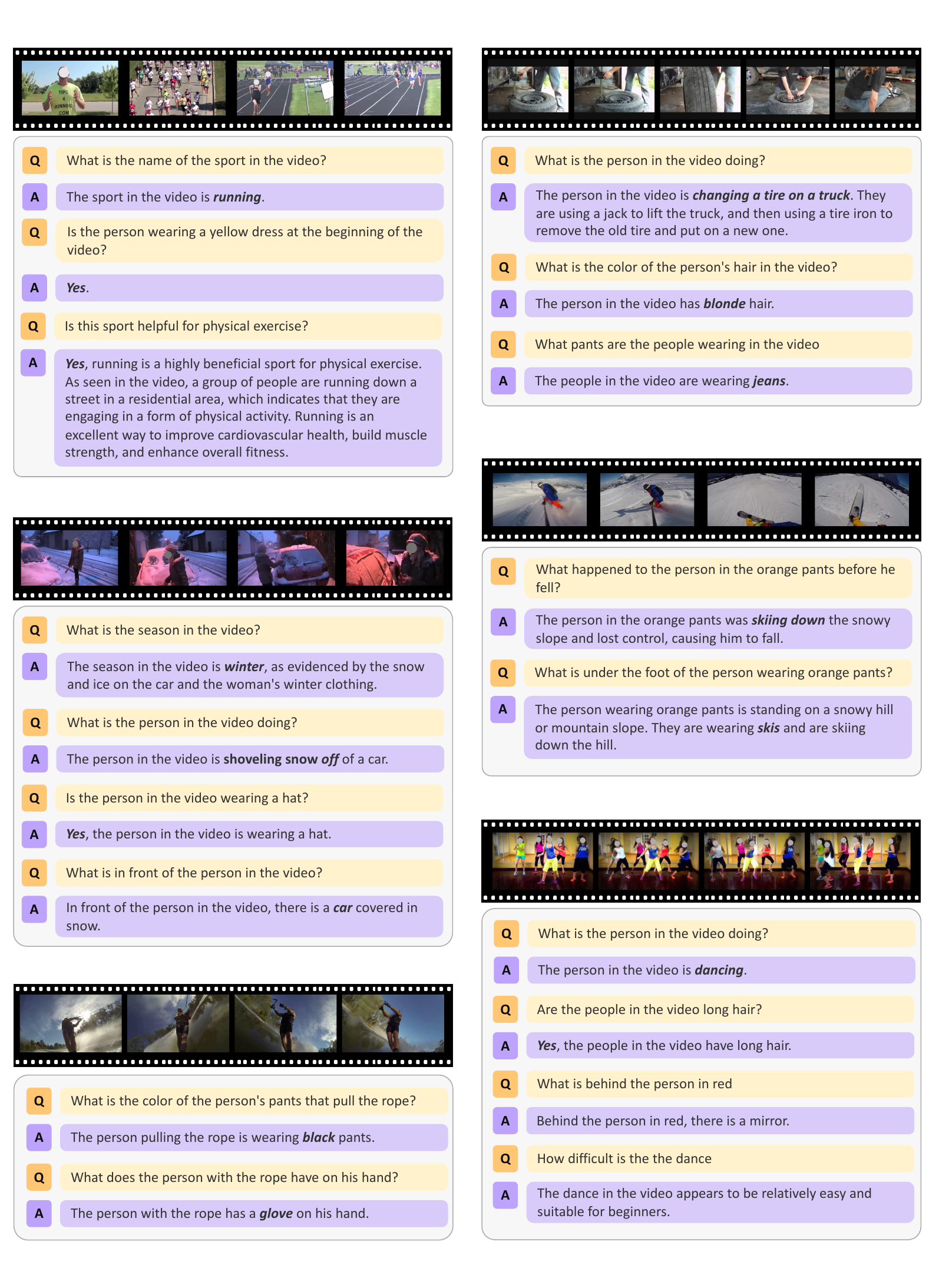}
  \caption{Additional examples from Zero-shot ANET-QA~\cite{yu2019activitynet} generated by the proposed \abbrmodel. Text in bold indicates the correct content. 
  }
  \label{fig:more_anet_qa}
\end{figure}

\begin{figure}[!t]
  \centering
  \includegraphics[width=1.0\linewidth]{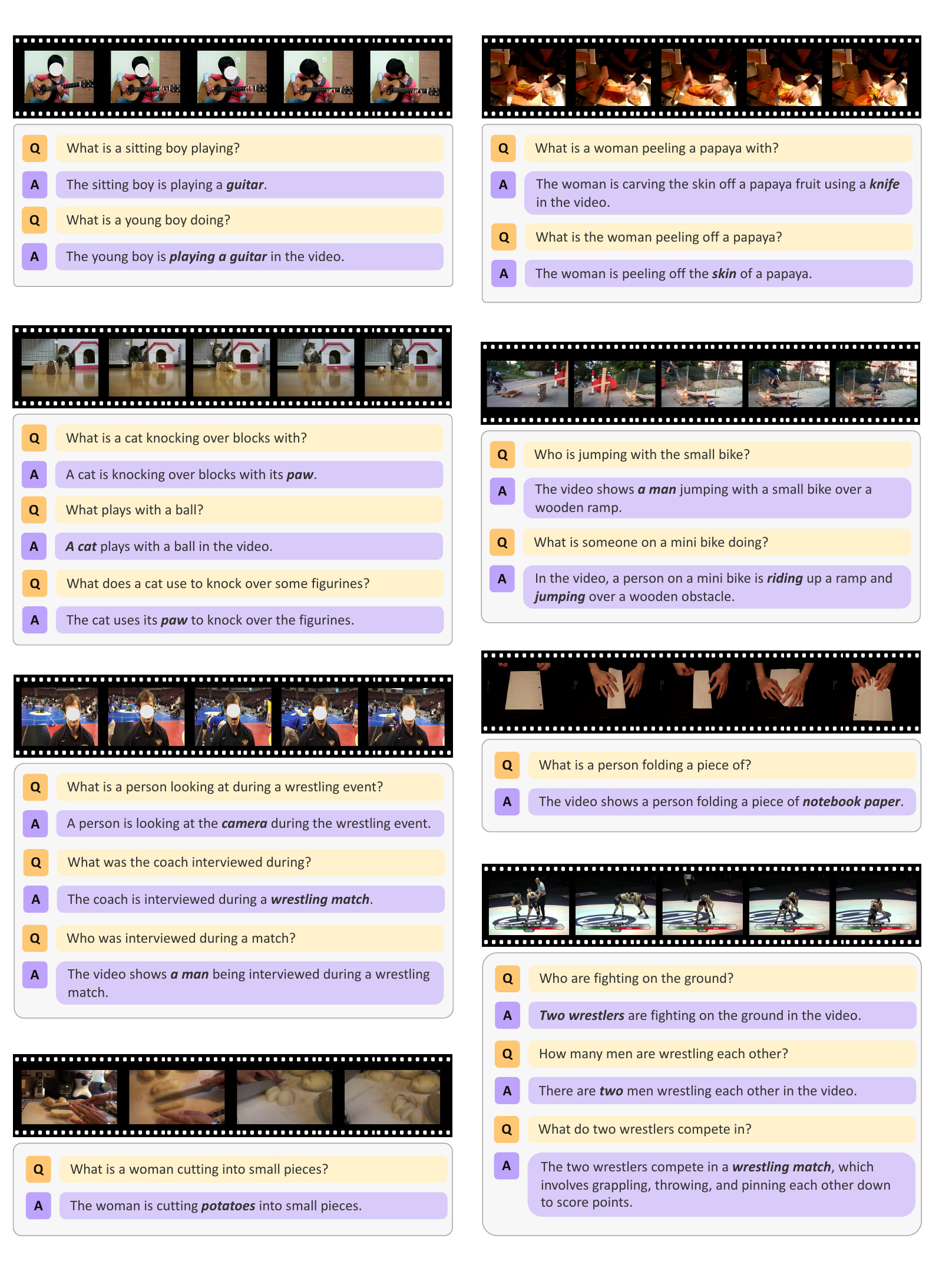}
  \caption{Response examples from Zero-shot MSVD-QA and MSRVTT-QA~\cite{xu2017video} generated by the proposed \abbrmodel. Text in bold indicates the correct content. 
  }
  \label{fig:more_msvd_qa}
\end{figure}

\end{document}